\documentclass[11pt,a4paper]{article}
\usepackage[hyperref]{conll-2019}
\usepackage{times}
\usepackage{latexsym}

\usepackage{url}

\usepackage{soul}
\usepackage[utf8]{inputenc}
\usepackage{graphicx}
\usepackage{amsmath}
\usepackage{booktabs}
\usepackage{algorithm}
\usepackage{algorithmic}
\usepackage{multirow}
\usepackage{array}

\makeatletter
\def\hlinew#1{%
    \noalign{\ifnum0=`}\fi\hrule \@height #1 \futurelet
    \reserved@a\@xhline}
\makeatother%\hlinewd{0.75pt}

\aclfinalcopy % Uncomment this line for the final submission

\title{Incorporating Interlocutor-Aware Context \\into Response Generation on Multi-Party Chatbots}

\author{
	Cao Liu$^{1,2}$,
	Kang Liu$^{1,2}$,
	Shizhu He$^{1,2}$,
	Zaiqing Nie$^3$, 
	Jun Zhao$^{1,2}$ \\
	$^1$ National Laboratory of Pattern Recognition, Institute of Automation, \\
	Chinese Academy of Sciences, Beijing, 100190, China \\
	$^2$ University of Chinese Academy of Sciences, Beijing, 100049, China \\
	$^3$ Alibaba AI Labs, Beijing, 100029, China \\
	\{cao.liu, kliu, shizhu.he, jzhao\}@nlpr.ia.ac.cn \\
	zaiqing.nzq@alibaba-inc.com
}

\date{}

\begin{document}
\maketitle
\begin{abstract}
    Conventional chatbots focus on two-party response generation, which simplifies the real dialogue scene. In this paper, we strive toward a novel task of Response Generation on Multi-Party Chatbot (RGMPC), where the generated responses heavily rely on the interlocutors' roles (e.g., speaker and addressee) and their utterances. Unfortunately, complex interactions among the interlocutors' roles make it challenging to precisely capture conversational contexts and interlocutors' information. Facing this challenge, we present a response generation model which incorporates Interlocutor-aware Contexts into Recurrent Encoder-Decoder frameworks (ICRED) for RGMPC. Specifically, we employ interactive representations to capture dialogue contexts for different interlocutors. Moreover, we leverage an addressee memory to enhance contextual interlocutor information for the target addressee. Finally, we construct a corpus for RGMPC based on an existing open-access dataset. Automatic and manual evaluations demonstrate that the ICRED remarkably outperforms strong baselines.
\end{abstract}

\section{Introduction}
\label{Setion: Introduction}
Human computer conversation has been an important and challenging task in NLP and AI since the Turing Test was proposed in 1950~\cite{utomated1950Computing}. Recently, with the rapid growth of social conversation data available on the Internet, data-driven chatbots are able to learn to generate responses directly and have attracted much more attention than before~\cite{li-EtAl:2016:N16-11,tian-EtAl:2017:Short}. % moradiannasab:2016:ACL-SRW,dhingra-EtAl:2017:Long1,

\begin{figure}[t]
    \begin{center}
        \includegraphics[width=215pt]{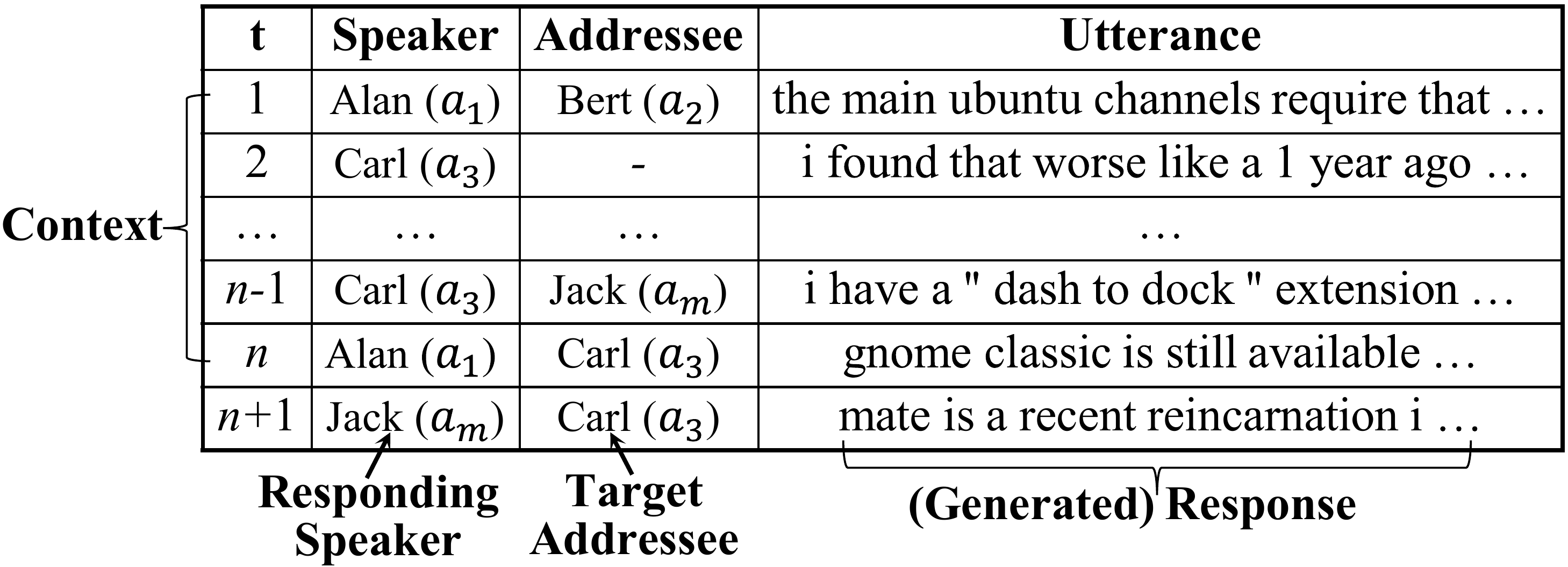}
        \caption{An example of Multi-Party Chatbots (MPC). At each turn, a speaker said one utterance to an addressee. There are many interlocutors (e.g., Alan, Bert and so on) in a conversation, where $a_i$ represents interlocutor's ID.}%     Response Generation on MPC (RGMPC) needs to generate responses at turn $n$+$1$ based on the dialog context, responding speaker ($a_3$) and target addressee ($a_{m}$).}
        \label{fig:example}
    \end{center}
\end{figure}

Researches in this area mostly focus on the dialog with two interlocutors~\cite{DBLP:journals/corr/BayserCSBCPB17}. However, the real-life interaction involves a substantial part of Multi-Party Chatbots (MPC, such as internet forum and chat group), which is a form of conversation with multiple interlocutors~\cite{ouchi-tsuboi:2016:EMNLP2016}. For example, there are more than three interlocutors ($a_1,a_2,a_3...a_{m}$) involved in the conversation in Figure \ref{fig:example}, and their roles (e.g., speaker and addressee) may change across different dialog turns. % meetings

As shown in Figure \ref{fig:example}, at each turn, the core issue of MPC is to capture \textit{who} (speaker) talks to \textit{whom} (addressee) about \textit{what} (utterance). In order to obtain responses in MPC, in our best knowledge, previous approaches usually employ a response selection paradigm, which simply selects one response from a set of existing utterances as the final response according to the contexts. Obviously, this paradigm, which could not generate new responses, is not so flexible. In this study, to build a more broadly applicable system, we concentrate on producing new responses word by word, named as Response Generation on Multi-Party Chatbots (RGMPC).

RGMPC is a very challenging task. The primary challenge is that the generated response has strong relevance to the interlocutor's roles, such as the speaker and the addressee. For example, in the same context of Figure \ref{fig:example}, what $a_{1}$ says to $a_{2}$ is different from what $a_{1}$ says to $a_{3}$ because different addressees ($a_{2}$ and $a_{3}$) have different information demands. Similarly, as for the same addressee, utterances from different speakers may be different because each speaker has personal background knowledge and style of speaking. Moreover, the roles of the same interlocutor may vary across different dialog turns. For instance, in Figure \ref{fig:example}, $a_{3}$ plays different roles in different dialog turns: speaker in the turn 2 and $n$-$1$, addressee in the turn $n$ and $n$+$1$.

Therefore, it is very important for RGMPC to capture interlocutor information. Currently, most response generation methods consider only the contextual utterance information \cite{Serban:2016:BED:3016387.3016435,serban2016hierarchical} but neglect the interlocutor information. Although some researches have exploited the interlocutor information for response generation, they are still suffering from certain critical limitations. \citeauthor{li-EtAl:2016:P16-13} \shortcite{li-EtAl:2016:P16-13} learn a fixed vector for each person from all conversational texts in the training corpus. However, as a global representation, the fixed person vector needs to be trained from large-scale dialogue turns for each interlocutor, and it may have a \textbf{sparsity issue} since some interlocutors have very few dialogue turns.%%\cite{ijcai2018-595}. % it is unable to handle \textbf{unknown persons} out of the training corpus. More importantly,

%In this paper, we first formalize the task of Response Generation on Multi-Party Chatbots (RGMPC), which is able to automatically generate responses based on the dialogue contexts, the responding speaker and the target addressee. 

To address the aforementioned problems of RGMPC, this paper incorporates Interlocutor-aware Contexts into a Recurrent Encoder-Decoder model (ICRED) for RGMPC, which is also an end-to-end framework. Specifically, in order to capture interlocutor information, we exploit interactive interlocutor representations learned from current dialog context rather than the fixed person vectors \cite{li-EtAl:2016:P16-13} obtained from all dialogs in the training corpus. We expect that the learned contextual interlocutor representation could be a good alternative to the fixed person vectors \cite{li-EtAl:2016:P16-13} due to its ability of alleviating the sparsity issue. Furthermore, from the view of conversation analysis, responses are usually used for answering the addressee's question or expanding the addressee's utterances. Therefore, we originally introduce an addressee memory mechanism to enhance contextual information for the target addressee especially. Finally, both of the interactive interlocutor representation and addressee memory are utilized for decoding response utterances. In particular, the addressee memory is leveraged to capture the addressee information for each generated word dynamically. % deal with unseen interlocutors out of the training corpus and

In order to prove the effectiveness of the proposed model, we construct a dataset for RGMPC based on an open dataset\footnote{The dataset is available at https://www.dropbox.com/s/4chh64yaxajh0j7/RGMPC.zip?dl=0\label{data_set}}. Experimental results show that the proposed model is fairly competitive on both automatic and manual evaluations compared with state-of-the-arts. % Our dataset is based on an open dataset (details in Section \ref{susetion: data}). Due to the maximum file size limit in the submission system, 

In brief, the main contributions of the paper are as follows:

%(1) We formalize a novel task of Response Generation on Multi-Party Chatbots (RGMPC), and it comes from a wide range of real-life conversation scenarios.

(1) We propose an end-to-end response generation model called ICRED which incorporates Interlocutor-aware Contexts into Recurrent Encoder-Decoder framework for RGMPC. % frameworks

(2) We leverage an addressee memory mechanism to enhance contextual interlocutor information for the addressee.

(3) We construct an open-access dataset for RGMPC. Both automatic and manual evaluations demonstrate that our model is remarkably better than strong baselines in this dataset.

\begin{figure*}[t]
    \begin{center}
        \includegraphics[width=450pt]{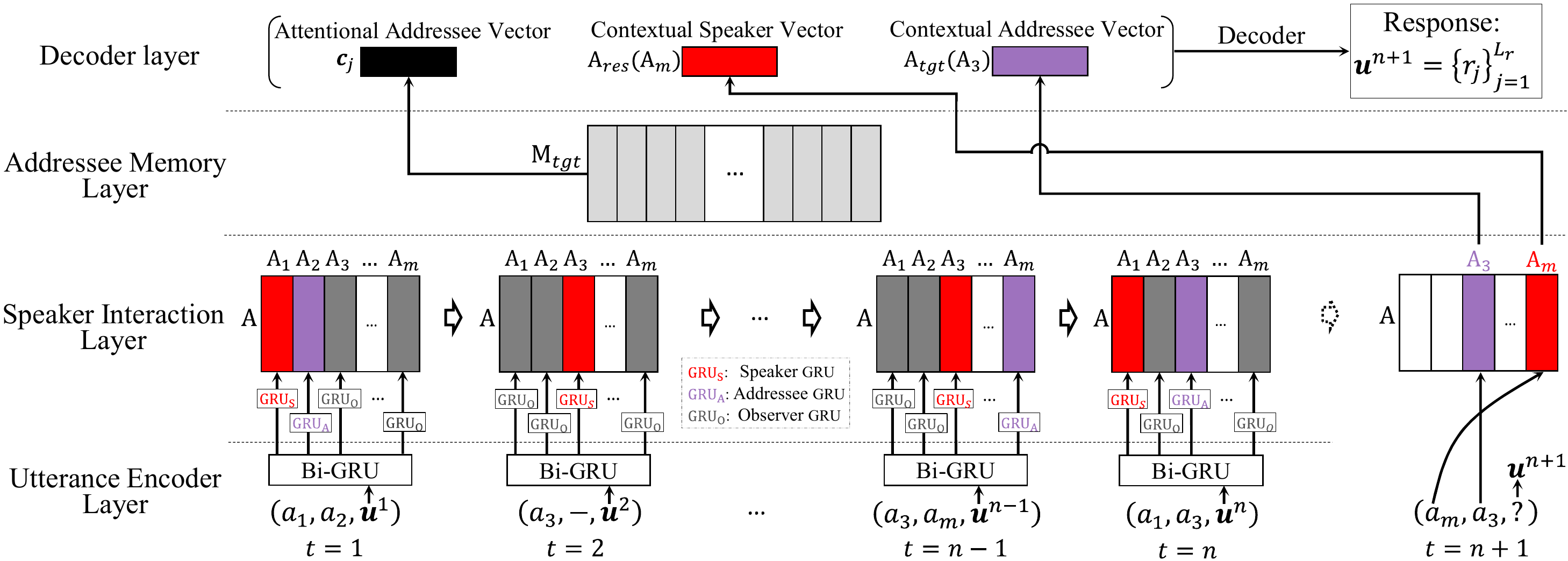}
        \caption{Overall structure of the proposed ICRED for RGMPC. At each time step $t$, $(a_{i},a_{j},\textbf{u}^{t})$ means that a speaker $a_{i}$ said an utterance $\textbf{u}^{t}$ to an addressee $a_{j}$, where the time step $t$ is denoted on the bottom, and the superscript $t$ may be omitted for brevity. Our ICRED includes: 
            \textcircled{1} Utterance Encoder Layer: encoding each utterance ($\textbf{u}^t$) into distributed vectors; \textcircled{2} Speaker Interaction Layer: capturing interactive interlocutor information from contexts, and it updates all interlocutors' representation by different GRUs according to their roles at each time step, where the embedding for an interlocutor $a_i$ is obtained by extracting the i-th column ($A_{i}$) from the interlocutor embedding matrix $A$; \textcircled{3} Addressee Memory Layer: enhancing contextual information for the target addressee ($a_3$); \textcircled{4} Decoder Layer: generating responses.} % : generating responses based on the interactive interlocutor representation and addressee memory.  % 1) Utterance Encoder Layer: encoding each utterance from input word sequence; 2) Speaker Interaction Layer: capturing interactive interlocutor information from contexts; 3) Addressee Memory Layer: enhancing contextual information for the target addressee; 4) Decoder Layer: generating responses
        \label{fig:overall-structure}
    \end{center}
\end{figure*}

\section{Task Formulation}
\label{setion:Task}
\iffalse
\begin{table}[h]\footnotesize%\footnotesize\scriptsize
    \centering
    \setlength{\tabcolsep}{4pt}
    \renewcommand\arraystretch{1.1}
    \begin{tabular}{ccc}
        \hlinew{1.2pt}
        & Data & Notation \tabularnewline
        \hline
        \multirow{3}{*}{Input} & Context & $\mathcal{C}$ \tabularnewline
        & Responding Speaker & $a_{spk}$ \tabularnewline
        & Responding Addressee & $a_{adr}$ \tabularnewline
        \hline
        Output & Response & $\textbf{r}$ \tabularnewline
        \iffalse
        \hline
        \multirow{3}{*}{ Context $\mathcal{C}$} & Speaker ID at time step $t$ & $a_{spk}^{t}$ \tabularnewline
        & Addressee ID at time step $t$ & $a_{adr}^{t}$ \tabularnewline
        & Utterance at time step $t$ & $\textbf{u}^{t}$ \tabularnewline
        \fi
        \hlinew{1.2pt}
    \end{tabular}
    \caption{Notations for RGMPC.}
    \label{tab:Notation}
\end{table}
\fi

\begin{table}[h]\footnotesize%\footnotesize\scriptsize
	\centering
	\setlength{\tabcolsep}{4pt}
	\renewcommand\arraystretch{1.1}
	\begin{tabular}{ccc}
		\hlinew{1.2pt}
		& Data & Notation \tabularnewline
		\hline
		\multirow{3}{*}{Input} & Context & $\mathcal{C}= [(a_{spk}^{t}, a_{adr}^{t}, \textbf{u}^{t})]_{t=1}^{n}$ \tabularnewline [3pt]
		& Responding Speaker & $a_{spk}^{n+1}$ 
		(or $a_{res}$) \\ [3pt]
		& Target Addressee & $a_{adr}^{n+1}$ (or $a_{tgt}$) \tabularnewline [1pt]
		\hline
		Output & Response & $\textbf{u}^{n+1}$ (or $\{r_j\}_{j=1}^{L_r}$) \tabularnewline
		\hlinew{1.2pt}
	\end{tabular}
	\caption{Notations for RGMPC.}
	\label{tab:Notation}
\end{table}

On multi-party chatbots, lots of interlocutors talk about one or more topics. At each dialogue turn (or time step) $t$, there is a speaker ($a_{spk}^{t}$), who may talk something ($\textbf{u}^{t}$) to a specific addressee ($a_{adr}^{t}$), while the others are observers. As shown in Table \ref{tab:Notation}, given the context $\mathcal{C}$ of previous $n$ dialog turns, the responding speaker $a_{res}$ and the target addressee $a_{tgt}$ at time step $n$+$1$, the task of RGMPC aims to automatically generate the next utterance $\textbf{u}^{n+1}$ as the final response. Here, $\mathcal{C}$ is a list ordered by the time step $t$: $\mathcal{C} = [\mathcal{C}^t]_{t=1}^{n} = [(a_{spk}^{t}, a_{adr}^{t}, \textbf{u}^{t})]_{t=1}^{n}$,
\iffalse
\begin{equation}\footnotesize
\begin{split}
\mathrm{G}iven: \quad &\mathcal{C} = [(a_{spk}^{t},\; a_{adr}^{t}, \textbf{u}^{t})]_{t=1}^{T} \\
& a_{spk} \\
& a_{adr} \\
\mathrm{P}redicate: \quad & \textbf{r}=\{w_{r_1},w_{r_2}...w_{r_{L_r}}\}
\end{split}
\label{eq:3411}
\end{equation}

\begin{equation}
\begin{split}
\mathcal{C} = [\mathcal{C}^t]_{t=1}^{n} = [(a_{spk}^{t}, a_{adr}^{t}, \textbf{u}^{t})]_{t=1}^{n}
\end{split}
\end{equation}
\fi
where $\mathcal{C}^t$ means $a_{spk}^{t}$ says $\textbf{u}^{t}$ to $a_{adr}^{t}$ at time step $t$, $n$ is the maximum number of previous dialog turns in a context. $\textbf{u}^{t}=(w_{1}^{t},w_{2}^{t}...w_{L_u}^{t})$ is the input utterance (word sequence) at time step $t$, where $L_u$ is the number of maximum words in utterances. %Table \ref{tab:Notation} shows the notations for RGMPC. 

\iffalse
\begin{CJK*}{UTF8}{song}
    \begin{table}[t]\footnotesize%\footnotesize\scriptsize
        \centering
        \setlength{\tabcolsep}{4pt}
        \renewcommand\arraystretch{1.1}
        \begin{tabular}{ccc}
            \hlinew{1.2pt}
            & 数据 & 标注 \tabularnewline
            \hline
            \multirow{3}{*}{输入} & 上下文 & $\mathcal{C}= [(a_{spk}^{t}, a_{adr}^{t}, \textbf{u}^{t})]_{t=1}^{n}$ \tabularnewline [3pt]
            & 响应说话者 & $a_{spk}^{n+1}$ 
            (or $a_{res}$) \\ [3pt]
            & 目标受话者 & $a_{adr}^{n+1}$ (or $a_{tgt}$) \tabularnewline [1pt]
            \hline
            输出 & (生成)回复 & $\textbf{u}^{n+1}$ (or $\{r_j\}_{j=1}^{L_r}$) \tabularnewline
            \hlinew{1.2pt}
        \end{tabular}
        \caption{Notations for RGMPC.}
        \label{tab:Notation}
    \end{table}
\end{CJK*}
\fi

\section{Methodology}
\label{Setion: Methodology}
The overview of the proposed ICRED for RGMPC is shown in Figure \ref{fig:overall-structure} along with its caption. The details are as follows. % The overall structure of the proposed ICRED for RGMPC is shown in Figure \ref{fig:overall-structure}, which is made up of 4 following submodules. 1) Utterance Encoder Layer; 2) Speaker Interaction Layer; 3) Addressee Memory Layer; 4) Decoder Layer.  1) Utterance Encoder Layer: encoding each utterance from input word sequence; 2) Speaker Interaction Layer: capturing interactive interlocutor information from contexts; 3) Addressee Memory Layer: enhancing contextual information for the target addressee; 4) Decoder Layer: generating responses

\subsection{Utterance Encoder Layer}
\label{subsetion: uttEnc}
The utterance encoder layer transforms input utterance into distributional representations. We leverage the bi-directional Gated Recurrent Units (GRU)~\cite{cho-EtAl:2014:EMNLP2014} to capture the long-term dependency. For an utterance $\textbf{u}^{t}=(w_{1}^{t},w_{2}^{t}...w_{L_u}^{t})$ at time step $t$, the concatenated representation for hidden states in bi-directions is denoted as
$\textbf{h}_{i}^{t}= [\overrightarrow{\textbf{h}}_{i}^{t},\overleftarrow{\textbf{h}}_{{L_u} - i + 1}^{t}]$
\iffalse
\begin{equation}
\begin{split}
\textbf{h}_{i}^{t}= [\overrightarrow{\textbf{h}}_{i}^{t},\overleftarrow{\textbf{h}}_{{L_u} - i + 1}^{t}]
\end{split}
\label{equ:ctx_word}
\end{equation}
\fi
, where $\textbf{h}_{i}^{t}$ is considered as the contextual word representation of the input word $w_{i}^{t}$. The state ($\textbf{h}_{L_u}^{t}$) of the last word is treated as the representation of the utterance at time step $t$, which is denoted as $\textbf{h}^{t}$, and it could be sent to the speaker interaction layer for updating contextual representation.

\subsection{Speaker Interaction Layer}
\label{subsetion: prsnEnc}
The speaker interaction layer is leveraged to obtain the interlocutor information in the context. Similar to the Speaker Interaction RNNs \cite{Zhang2017Addressee}, we utilize the interactive speaker encoder for RGMPC.

As shown in Figure \ref{fig:overall-structure}, an interlocutor embedding matrix $A$ is used to record all interlocutors' representation, and $A$ is initiated with a zero matrix. Each column of $A$ corresponds to an interlocutor's embedding: $A_{i}=A[*, a_{i}]$, where $A_{i}$ is the embedding for the interlocutor $a_{i}$. The speaker interaction layer updates the entire interlocutors' embeddings at each time step based on their roles (speaker, addressee or observer). Embeddings for the speaker, addressee and observer are updated by following role-differentiated GRUs: $\mathrm{GRU}_{S}$, $\mathrm{GRU}_{A}$ and $\mathrm{GRU}_{O}$, respectively.% For ``$(a_{spk}, a_{adr}, \textbf{u}^t)$"\footnote{The complete form is $(a_{spk}^{t}, a_{adr}^{t}, \textbf{u}^t)$. Sometimes, the superscript $t$ is omitted for sake of brevity.} at time step $t$, the speaker's embedding $A_{spk}$ is updated through $\mathrm{GRU}_{S}$:  Similarly, embeddings for the addressee and observers are updated by $\mathrm{GRU}_{A}$ and $\mathrm{GRU}_{O}$, respectively.
%{\footnotesize
\begin{gather}
A_{spk}^{t}=\mathrm{GRU}_{S}(A_{spk}^{t-1}, \textbf{h}^{t}) \\
A_{adr}^{t}=\mathrm{GRU}_{A}(A_{adr}^{t-1}, \textbf{h}^{t}) \\
A_{obv}^{t}=\mathrm{GRU}_{O}(A_{obv}^{t-1}, \textbf{h}^{t})
\end{gather}
%    }
% $$A_{spk}^{t}=\mathrm{GRU}_{S}(\textbf{a}_{spk}^{t-1}, \textbf{h}^{t})$$
where $A_{spk}^{t}$ ($A_{adr}^{t}$ / $A_{obv}^{t}$) is the embedding for the speaker (addressee / observer) at time step $t$, and $\textbf{h}^{t}$ is the utterance representation obtained from the utterance encoder layer. Take the first time step ``$(a_{1},a_{2},\textbf{u}^{1})$" in Figure \ref{fig:overall-structure} as an example, when $a_{1}$ says $\textbf{u}^{1}$ to $a_{2}$, the speaker's ($a_{1}$'s) embedding $A_{1}$ is updated by the speaker GRU---$\mathrm{GRU}_{S}$, and the addressee's ($a_{2}$'s) embedding $A_{2}$ is updated by the addressee GRU---$\mathrm{GRU}_{A}$, while other interlocutors' embeddings are updated by the observer GRU---$\mathrm{GRU}_{O}$. Note that the addressee may be missing (such as ``$(a_{3},-,\textbf{u}^{2})$" at time step 2 in Figure \ref{fig:overall-structure}), where embeddings for all interlocutors except for the speaker are updated by the observer GRU. The interlocutor embedding matrix ($A$) is updated up to the maximum time step $n$. The final interlocutor embedding matrix is used in decoding.

\subsection{Addressee Memory Layer}
\label{Subsetion: Addressee_Memory}
The interlocutor embedding matrix is updated by utterance representations and interlocutor's roles, so it captures interlocutor's context on the utterance level. In fact, contextual word representation is important for response generation, too. A context contains consecutive utterances, and each utterance is a word sequence. Therefore, memorizing all contextual word representations in the entire context is complex, and it is difficult to work on large-scale utterances in one context.

Intuitively, from the view of conversational analysis, responses are usually used for answering the addressee's question or expanding the addressee's utterances. Therefore, we design an addressee memory layer, which only memorizes the contextual word representations (noted as $M_{tgt}$) in the last utterance said by the target addressee, and the contextual representation for each word is obtained from the utterance encoder layer. Take ``$(a_m,a_3,?)$" at time step $n$+$1$ in Figure \ref{fig:overall-structure} as an example, $\textbf{u}^{n-1}$ is the last utterance said by the target addressee $a_3$ because of ``$(a_3,a_m,\textbf{u}^{n-1})$" at time step $n$-$1$, so the addressee memory layer merely memorizes contextual word representation $M_{tgt} = [\textbf{h}_{1}^{n-1},\textbf{h}_{2}^{n-1},...,\textbf{h}_{L_u}^{n-1}]$ from the utterance $\textbf{u}^{n-1}$, where $\textbf{h}_{i}^{n-1}$ is obtained from Section \ref{subsetion: uttEnc}.

\subsection{Decoder Layer}
\label{Subsetion: dec}
The decoder is responsible for generating target sequences. Different from a single contextual representation in previous work \cite{serban2016hierarchical}, the speaker interaction layer is able to capture different interlocutor information from contexts (e.g., personal background knowledge and style of speaking for the responding speaker, special information demands for the target addressee). Moreover, the addressee memory layer records contextual word representation for the target addressee. Therefore, we extract \textbf{contextual speaker vector} $A_{res}$ for the responding speaker $a_{res}$ from the final interlocutor embedding matrix $A$ (e.g., the responding speaker's embedding obtained by $A_m=A[*, a_m]$ for the responding speaker $a_m$ in Figure \ref{fig:overall-structure}). Similarly, \textbf{contextual addressee vector} $A_{tgt}$ for the target addressee is also extracted from $A$. However, $A_{res}$ and $A_{tgt}$ keep same for each generated word. In order to capture dynamic information for different generated words, we leverage an attention mechanism to selectively reads different contextual word representations from the addressee memory. For each target word, the decoder attentively reads the contextual word representation as follows:
\begin{gather}%\footnotesize
\textbf{c}_j = \sum \nolimits_{k=1} \nolimits^{L_u}\alpha_{jk}M_{tgt}[*, k]; \label{eq:attention0} \\
\alpha_{jk}=\frac{e^{\rho(\textbf{s}_{j-1},M_{tgt}[*, k])}}{\sum\nolimits_{k'}e^{\rho(\textbf{s}_{j-1},M_{tgt}[*, k'])}}
\label{eq:attention}
\end{gather}
where $\textbf{c}_j$ is the \textbf{attentional addressee vector}, $M_{tgt}[*, k]$ is the contextual word representation for the k-th word in the addressee memory, and $\textbf{s}_j$ represents the hidden state in decoding GRU. A function $\rho$ is leveraged to compute the attentive strength, which is calculated by a projected matrix to connect $\textbf{s}_{j-1}^T$ and $M_{tgt}[*, k]$. Finally, the attentional addressee vector $\textbf{c}_j$, contextual speaker vector $A_{res}$ and contextual addressee vector $A_{tgt}$ are concatenated to estimate the probability for predicted words:
\begin{gather}%\footnotesize
\begin{split}
p(r_j|r_{<j}, a_{spk}, a_{tgt}, \mathcal{C}) = ~~\\
~~~~p(r_{j}|r_{j-1}, \textbf{c}_j, A_{res}, A_{tgt}, \textbf{s}_j)
 \label{eq:pred}
\end{split} \\
\textbf{s}_j=\mathrm{GRU}_{dec}(\textbf{s}_{j-1}, [\textbf{c}_j, A_{res}, A_{tgt}, \textbf{x}_{j-1}])
 \label{eq:312}
\end{gather}%     }
where $\textbf{s}_j$ is the hidden state of the decoding GRU---${\rm GRU_{dec}}$. $\textbf{x}_j$ is the word vector of the predicted target word $r_j$, and $r_j$ is typically performed by a $softmax$ classifier over a settled vocabulary based on word embedding similarity.

\subsection{Learning}
\label{Subsetion: learning}
The proposed ICRED for RGMPC is totally differentiable, and it can be optimized in an end-to-end manner using back-propagation. Given the context $\mathcal{C}$, responding speaker $a_{res}$, target addressee $a_{tgt}$ and target word sequence $\{r_{j}\}_{j=1}^{L_r}$, the objective function is to minimize the loss function:
%\vspace{-8.6pt}
\begin{equation}%\footnotesize
\begin{split}
%\mathcal{L} = \mathcal{L}_{r} + \frac{\lambda}{2}||\theta||^2 ~~~~~~~~~~~~~~~~~~~~~~ \\
%\mathcal{L}_{r} = -\frac{1}{L_r}\sum_{j=1}^{L_r}log[p(r_j|r_{<j},a_{spk}, a_{adr},\mathcal{C}]
\mathcal{L} = \frac{-1}{L_r}\sum_{j=1}^{L_r}log[p(r_j|r_{<j},\mathcal{C}, a_{res}, a_{tgt}] + \lambda \mathcal{L}_2 % \nolimits
\end{split}
\label{eq:401}
\end{equation}

It contains a negative log-likelihood for generated responses and L2 regularization ($\mathcal{L}_2$), where $\lambda$ is a hyperparameter for $\mathcal{L}_2$. % for the generated response

\section{Experiment}
\label{Setion: Experiment}

%\subsection{Dataset and Training Details}
\subsection{Dataset}
\label{susetion: data}
\begin{table}[h]\footnotesize%\footnotesize\scriptsize
	\centering
	\setlength{\tabcolsep}{4pt}
	\renewcommand\arraystretch{1.1}
	%\begin{tabular}{ccccc}
	\begin{tabular}{ccccc}
		\hlinew{1.2pt}
		& \textbf{Total} & \textbf{Train} & \textbf{Dev} & \textbf{Test} \tabularnewline
		\hline
		\# Contexts & 423.5K & 338.9K & 42.3K & 42.3K \tabularnewline
		\# Speaker & 35.3K & 33.5K & 15.6K & 15.6K \tabularnewline
		\# Addressee & 23.4K & 22.4K & 10.8K & 10.8K \tabularnewline
		% Avg. Spk/Ctx & 2.99 & 2.99 & 2.99 & 3.00 \tabularnewline
		% Avg. Adr/Ctx & 1.83 & 1.83 & 1.83 & 1.84 \tabularnewline
		\# Vocab & 276.1K & 254.8K & 82.2K & 82.0K \tabularnewline
		\# Tokens & 26.3M & 21.0M & 2.62M & 2.62M \tabularnewline
		Avg. Tok/Ctx & 51.4 & 51.5 & 51.4 & 51.3 \tabularnewline
		Avg. Tok/Res & 10.6 & 10.6 & 10.7 & 10.6 \tabularnewline
		\hlinew{1.2pt}
	\end{tabular}
	\caption{Data statistics. ``\#" means number, and ``Avg. Tok/Ctx (or Res)" is the number of tokens per context (or response).} % ``Avg. Spk/Ctx" and ``Avg. Adr/Ctx" are the average number of different speakers and addressees for a context, respectively.
	\label{tab: data}
\end{table}

Our dataset is constructed based on the Ubuntu multi-party chatbot corpus\footnote{https://github.com/hiroki13/response-ranking}, which has been widely used as the evaluation dataset for the response selection task~\cite{ouchi-tsuboi:2016:EMNLP2016,Zhang2017Addressee}. The original data comes from the Ubuntu IRC chat log, where each line consists of (Time, Speaker, Utterance). If the addressee is explicitly mentioned in the utterance, it is extracted as the addressee. Otherwise, all interlocutors except the speaker are observers.
Considering that generating new responses in this paper is more complicated than retrieving responses, the generative task requires higher-quality data. We suppose that the responding speaker and target addressee have appeared in the context, where the contextual window is set to 5. Moreover, the words are tokenized by NLTK, and some general responses are removed by human rules\footnote{We list some general responses, such as containing ``\textit{i don't know}", ``\textit{you are welcome}".}. Finally, we randomly split the dataset into Train/Dev/Test (8:1:1), and it is publicly available\textsuperscript{\ref{data_set}}. The detailed statistics of the
dataset are shown in Table \ref{tab: data}.

%In order to keep our model comparable to other typical existing methods, we keep the same parameters and experimental environments for ICRED and the comparative models. We take a maximum of 20 words for the utterance. The word vector dimension is 300 and it is initialized with the public released fasttext\footnote{https://github.com/facebookresearch/fastText} pre-trained on Wikipedia. The utterance and interlocutor are encoded by 512-dimensional and 1024-dimensional vectors, respectively. The joint loss function with 0.0001 L2 weight is minimized by an Adam optimizer. We implemented all the models with Tensorflow on an NVIDIA TITAN X GPU. %  \cite{kingma2014adam} % the open-sourced % \footnote{https://www.wikipedia.org/}

\subsection{Implement Details}
In order to keep our model comparable to other typical existing methods, we keep the same parameters and experimental environments for ICRED and the comparative models. We take a maximum of 20 words for the utterance. The word vector dimension is 300 and it is initialized with the public released fasttext\footnote{https://github.com/facebookresearch/fastText} pre-trained on Wikipedia. The utterance and interlocutor are encoded by 512-dimensional and 1024-dimensional vectors, respectively. The joint loss function with 0.0001 L2 weight is minimized by an Adam optimizer. We implemented all the models with Tensorflow on an NVIDIA TITAN X GPU.

\subsection{Automatic Evaluation Metrics}
Automatic evaluations (AEs) for Natural Language Generation (NLG) is a challenging and under-researched problem \cite{novikova-EtAl:2017:EMNLP2017}. Following \cite{ijcai2018-587}, we leverage two referenced measurements (BLEU~\cite{papineni-EtAl:2002:ACL} and ROUGE~\cite{Lin:2004}\footnote{Implemented by https://github.com/Maluuba-/nlg-eval. BLEU and ROUGE are transformed into percentages (\%).}) for automatic evaluations. Considering that current data-driven approaches tend to generate short and generic (meaningless) responses, two unreferenced (``intrinsic") metrics are also leveraged to the evaluation. The first one is the average length of responses, which is an objective and surfaced metric reflected the substance of responses~\cite{mou-EtAl:2016:COLING,he-EtAl:2017:Long4}. The other one is the number of nouns\footnote{NLTK is utilized for part-of-speech tagging.} per response \cite{ijcai2018-587}, which shows the richness of responses since nouns are usually content words. Note that the unreferenced metrics could enrich the evaluations, though they are weak metrics. The detailed results and analyses are shown as follows.

\subsection{The Effectiveness of ICRED for RGMPC}
\begin{table}[h]\footnotesize%\footnotesize\scriptsize
    \centering
    \setlength{\tabcolsep}{4pt}
    \renewcommand\arraystretch{1.1}
    \begin{tabular}{p{2.2cm}<{\centering}p{1.0cm}<{\centering}p{1.10cm}<{\centering}p{1.0cm}<{\centering}p{1.0cm}<{\centering}}
        %\begin{tabular}{|c|c|c|c|c|}
        \hlinew{1.2pt}
        \multirow{2}{*}{\textbf{Model}} & \multicolumn{2}{c}{\textbf{Referenced}} & \multicolumn{2}{c}{\textbf{Unreferenced}}
        \tabularnewline
        \cline{2-5}
        & \textbf{BLEU} & \textbf{ROUGE} & \textbf{Length} & \textbf{\#Noun} \tabularnewline
        \hline
        Seq2Seq & 8.86 & 7.62 & 9.48  & 1.24 \tabularnewline
        Persona Model & 9.12 & 7.38 & 11.04  & 1.29 \tabularnewline
        VHRED & 9.38 & 7.65 & 10.25  & 1.55 \tabularnewline \hline
        \textbf{ICRED} (ours) & \textbf{10.63} & \textbf{8.73} & \textbf{11.34}  & \textbf{1.68} \tabularnewline 
        \hlinew{1.2pt}
    \end{tabular}
    \caption{Overall comparisons of ICRED.}
    \label{tab: Overall Performances}
\end{table}
\textbf{Comparison Methods}. We compared ICRED with the following methods:

(1) Seq2Seq \cite{sutskever2014sequence}: Seq2Seq is one of the mainstream methods for text generation. In order to capture as much information as possible, the input sequence is all utterances concatenated in order in a context. %% , and it serves as a baseline

(2) Persona Model \cite{li-EtAl:2016:P16-13}: The persona-based model modified a Seq2Seq to encode a global vector for each interlocutor that appears in the training data, and it could alleviate the issue of speaker consistency for response generation.

(3) VHRED \cite{serban2016hierarchical}:  VHRED is essentially a conditional variational auto-encoder with hierarchical encoders, and it extends HRED \cite{Serban:2016:BED:3016387.3016435} by adding a high-dimensional latent variable for utterances.

\textbf{Comparative Results.}
Table \ref{tab: Overall Performances} demonstrates overall comparisons of ICRED. We can clearly obtain the following observations:

(1) ICRED obtains the highest performance on all metrics (marked as \textbf{bold}), and it indicates that incorporating interlocutor-aware context into RGMPC contributes to generating better responses.

(2) Although the persona-based model utilizes interlocutor information, it performs poorly. The average dialogue turn for the interlocutor is more than 5000 in \cite{li-EtAl:2016:N16-11}, while there is less than 100 dialogue turns per interlocutor in our dataset. Therefore, it is hard to learn a global vector for each interlocutor from the sparse corpus. In contrast, our ICRED performs well on such a sparse corpus (details in Section \ref{susetion: Sparse Data}).

(3) VHRED brings slight improvements over the Seq2Seq and persona-base model. Even that VHRED enhances the contextual information by a high-dimensional latent variable, VHRED is still remarkably worse than ICRED because VHRED neglects the interlocutor information.

%\subsection{The Effectiveness of ICRED on Sparse Data}
\subsection{The Effect of Sparse Data on ICRED}
\label{susetion: Sparse Data}
\begin{table}[t]\footnotesize%\footnotesize\scriptsize
    \centering
    \setlength{\tabcolsep}{4pt}
    \renewcommand\arraystretch{1.1}
    \begin{tabular}{p{2.1cm}<{\centering}p{0.95cm}<{\centering}p{1.05cm}<{\centering}p{1.05cm}<{\centering}p{1.05cm}<{\centering}}
        %\begin{tabular}{|c|c|c|c|c|}
        \hlinew{1.2pt}
        \textbf{Interlocutor's} & \multicolumn{2}{c}{\textbf{Persona Model}} & \multicolumn{2}{c}{\textbf{ICRED (ours)}}
        \tabularnewline
        \cline{2-5}
        \textbf{Dialogue Turns} & \textbf{~BLEU} & \textbf{ROUGE} & \textbf{BLEU} & \textbf{ROUGE} \tabularnewline
        \hline
        [0, 100] & 8.47 & 6.72 & 10.63 & 8.60 \\
        (100, 1000] & 8.87 & 7.14 & 10.50 & 8.61 \\
        (1000, 5000] & 9.48 & 7.74 & \textbf{10.77} & \textbf{8.90} \\
        (5000, $\;$+$\infty$) & \textbf{9.51} & \textbf{7.80} & 10.60 & 8.79 \\
        \hlinew{1.2pt}
    \end{tabular}
    \caption{Performances on sparse and plentiful learning data with different numbers of interlocutor's dialogue turns, where the test data is divided into different intervals according to the number of dialogue turns in training dataset said by target addressee (named as interlocutor's dialogue turns).}
    \label{tab: turns}
\end{table}
\textbf{Comparison Settings}. Persona model \cite{li-EtAl:2016:P16-13} may have a sparsity issue since some interlocutors have very few dialogue turns. To investigate whether ICRED has the sparsity issue or not, we divide the test data into four intervals according to the number of training dialogue turns said by the target addressee (called interlocutor dialogue turns), where small turns represent sparse learning data (e.g., ``[0, 100]") and large turns mean plentiful learning data (e.g., ``(5000, +$\infty$)").

\textbf{Comparative Results.} Table \ref{tab: turns} reports the performances of persona model and ICRED on different interlocutor's dialogue turns for learning. We can clearly see that the persona model has a sparsity issue: it performs very poorly on sparse learning data (e.g., BLEU score = 8.47 on ``[0, 100]") while it achieves good performances on plentiful learning data (e.g., BLEU score = 9.51 on ``(5000, +$\infty$)"), which demonstrates that the fixed person vectors in the persona model need to be learned from large-scale training data for each interlocutor. In contrast, ICRED exploits interactive interlocutor representation learned from current dialog context rather than the fixed person vectors obtained from all training dialog utterances. Therefore, ICRED has no sparsity issues and it performs closely on sparse and plentiful learning data.

\begin{figure*}[t]
    \begin{center}
        \includegraphics[width=450pt]{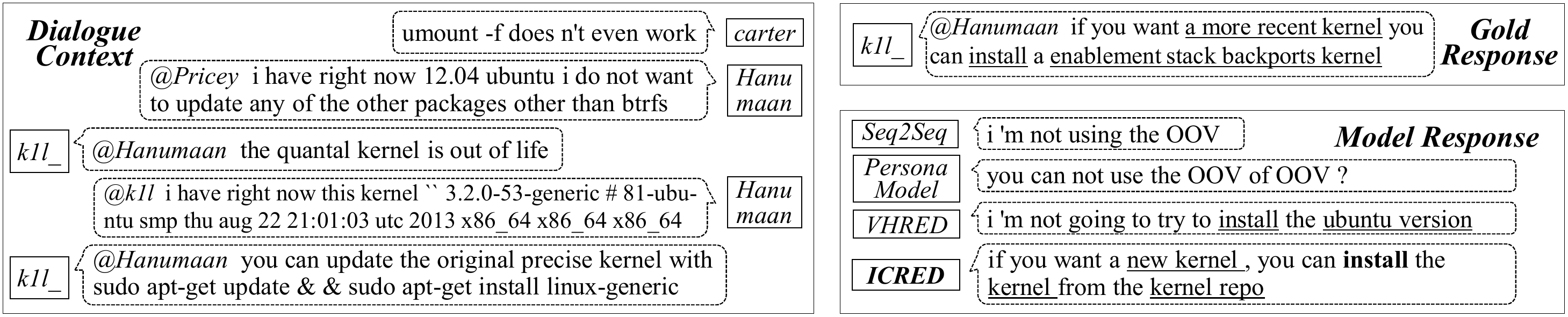}
        \caption{An example of different model responses for the same dialogue context. The input dialogue context is on the left. The gold (referenced) response and model responses are on the top right and bottom right, respectively. The rounded rectangle is the message box, where the \textit{italic} behind ``@" is the addressee, and the solid-line box near to the message box represents the speaker or model.}
        \label{fig:case}
    \end{center}
\end{figure*}

\subsection{Ablation Study for Model Components}
\begin{table}[h]\footnotesize%\footnotesize\scriptsize
    \centering
    \setlength{\tabcolsep}{4pt}
    \renewcommand\arraystretch{1.1}
    \begin{tabular}{p{2.2cm}<{\centering}p{1.0cm}<{\centering}p{1.10cm}<{\centering}p{1.0cm}<{\centering}p{1.0cm}<{\centering}}
        %\begin{tabular}{|c|c|c|c|c|}
        \hlinew{1.2pt}
        \multirow{2}{*}{\textbf{Model}} & \multicolumn{2}{c}{\textbf{Referenced}} & \multicolumn{2}{c}{\textbf{Unreferenced}}
        \tabularnewline
        \cline{2-5}
        & \textbf{BLEU} & \textbf{ROUGE} & \textbf{Length} & \textbf{\#Noun} \tabularnewline
        \hline
        \textbf{ICRED} & \textbf{10.63} & \textbf{8.73} & \textbf{11.34}  & \textbf{1.68} \tabularnewline \hline
        w/o Adr\_Mem & 10.25 & 8.23 & 10.73  & 1.27 \tabularnewline % Att\_Adr\_Vec
        w/o Ctx\_Spk\_Vec & 10.13 & 8.22 & 10.86  & 1.59 \tabularnewline
        w/o Ctx\_Adr\_Vec & 9.95 & 8.18 & 10.93  & 1.26 \tabularnewline
        \hlinew{1.2pt}
    \end{tabular}
    \caption{Ablation Experiments by removing the main components.} %%  in the decoder
    \label{tab: Ensemble Model}
\end{table}
\textbf{Comparison Settings}. In order to validate the effectiveness of model components, we have tried to remove some main components in decoding as follows. (1) w/o Adr\_Mem: without the addressee memory, such as removing $\textbf{c}_j$ in Equation \ref{eq:pred}-\ref{eq:312}; (2) w/o Ctx\_Spk\_Vec: without the contextual speaker vector, such as removing $A_{res}$ in Equation \ref{eq:pred}-\ref{eq:312}; (3) w/o Ctx\_Adr\_Vec: without the contextual addressee vector, such as removing $A_{tgt}$ in Equation \ref{eq:pred}-\ref{eq:312}. % \ref{eq:pred}-

\textbf{Comparative Results.}
Results of the ablation study are shown in Table~\ref{tab: Ensemble Model}. We can see that removing any component causes obvious performance degradation. In particular, ``w/o Ctx\_Adr\_Vec" performs the worst on almost all of the metrics, which demonstrates the importance of contextual information for the target addressee.

\subsection{The Effectiveness of Addressee Memory}
\begin{table}[h]\footnotesize%\footnotesize\scriptsize
    \centering
    \setlength{\tabcolsep}{4pt}
    \renewcommand\arraystretch{1.1}
    \begin{tabular}{p{2.68cm}<{\centering}p{0.8cm}<{\centering}p{1.0cm}<{\centering}p{1.0cm}<{\centering}p{0.8cm}<{\centering}}
        %\begin{tabular}{|c|c|c|c|c|}
        \hlinew{1.2pt}
        \multirow{2}{*}{\textbf{Memory Type}} & \multicolumn{2}{c}{\textbf{Referenced}} & \multicolumn{2}{c}{\textbf{Unreferenced}}
        \tabularnewline
        \cline{2-5}
        & \textbf{BLEU} & \textbf{ROUGE} & \textbf{Length} & \textbf{\#Noun} \tabularnewline
        \hline
        \textbf{addressee memory} & \textbf{10.63} & 8.73 & 11.34 & \textbf{1.68} \tabularnewline \hline
        all utterance memory & 10.39 & \textbf{8.78} & \textbf{11.38} & 1.37 \tabularnewline
        latest memory & 10.43 & 8.40 & 10.16 & 1.28 \tabularnewline
        speaker memory & 10.03 & 8.28 & 10.72 & 1.66 \tabularnewline
        w/o
        memory & 10.25 & 8.23 & 10.73 & 1.27 \tabularnewline
        \hlinew{1.2pt}
    \end{tabular}
    \caption{Performances over different memory types.}
    \label{tab: memory}
\end{table}

\textbf{Comparison Settings.} In order to demonstrate the effectiveness of the addressee memory, we change the memory type, and then the attention model in Equation \ref{eq:attention} is based on the new memory. The comparison settings are shown as follows. (1) addressee memory: memorizing contextual word representations in the last utterance said by the \textit{target addressee} (e.g., $\textbf{u}^{n-1}$ in Figure \ref{fig:overall-structure}); (2) all utterance memory: memorizing contextual word representations in \textit{all} utterances of the context (e.g., $\textbf{u}^{1}$ to $\textbf{u}^{n}$ in Figure \ref{fig:overall-structure}); (3) latest memory: memorizing contextual word representations of the \textit{latest} utterance in the context (e.g., the latest utterance $\textbf{u}^{n}$ in Figure \ref{fig:overall-structure}); (4) speaker memory: memorizing contextual word representations in the last utterance said by the \textit{responding speaker}; (5) w/o memory: without any memory.%don't memorize any contextual word representations.

\textbf{Comparative Results.} We report the results of different memory types as shown in Table \ref{tab: memory}. It can see that our method, the addressee memory, achieves the best or near-best performances on all metrics. Although memorizing all utterances is competitive, the complexity of all utterance memory is $n$ times compared with the one in the addressee memory, where $n$ is the number of utterances in a context. The speaker memory performs closely to without memory, which indicates that not all memories can improve the performance.

\subsection{Manual Evaluations}
Besides automatic evaluations, we employ manual evaluations (MEs), which is important for response generation. Similar to \cite{he-EtAl:2017:Long1,ijcai2018-643}, and we select three metrics for MEs, which measure the following aspects. (1) Fluency: measuring whether responses are grammatically correct or wrong. (2) Consistency: measuring whether responses are coherent to the context or not. (3) Informativeness: measuring how much informational (knowledgeable) content obtained from the responses.

\begin{table}[h]\footnotesize%\footnotesize\scriptsize
    \centering
    \setlength{\tabcolsep}{4pt}
    \renewcommand\arraystretch{1.1}
    \begin{tabular}{p{2.8cm}<{\centering}p{1.10cm}<{\centering}p{1.10cm}<{\centering}p{1.10cm}<{\centering}<{\centering}}
        %\begin{tabular}{|c|c|c|c|c|}
        \hlinew{1.2pt}
        
        \textbf{Model} & \textbf{Flu.} & \textbf{Con.} & \textbf{Inf.} \tabularnewline
        \hline
        ICRED vs. Seq2Seq & 77.25 & 83.69 & 84.35 \tabularnewline
        ICRED vs. Persona. & 78.44 & 80.41 & 82.35 \tabularnewline
        ICRED vs. VHRED & 73.20 & 81.29 & 79.47 \tabularnewline
        \hlinew{1.2pt}
    \end{tabular}
    \caption{Manual evaluations (\%) with fluency (Flu.), consistency (Con.) and informativeness (Inf.). The Score is the percentage that ICRED wins baselines after removing the ``tie" pairs.}% ICRED significantly outperforms (sign test, p-value $<$ 0.005) the competitors on all MEs.}
    \label{tab: ME}
\end{table}

We conduct a pair-wise comparison between the response generated by ICRED and the one for the same input by three typical baselines. We sample 100 responses from each compared methods. Two curators judge (win, tie and lose) between these two methods. The Cohen Kappa of inter-annotator statistics is 0.750, 0.658 and 0.580 for the fluency, consistency and informativeness, respectively. As shown in Table~\ref{tab: ME}, the score is the percentage that ICRED wins baselines after removing the ``tie" pairs, and we can obtain that ICRED is significantly (sign test, p-value $<$ 0.005) superior to all baselines on any metric. It demonstrates our model is able to deliver more fluent, consistent and informative responses. % in a very high proportion 

\subsection{Case Study}

Figure~\ref{fig:case} shows an example of responses on different models for the same dialogue context. It is clearly observed that our model (\textbf{ICRED}) generates more fluent, consistent and knowledgeable (marked as \underline{underline}) responses compared to baselines. In particular, the response given by ICRED ``\textit{if you want a new kernel , you can install the kernel from the kernel repo}", not only explains the \textit{reason for kernel installation} but also suggests a \textit{source of the installation}. It fully captures the context and then produces a fluent, consistent and knowledgeable response, which is semantically similar to the gold one.

\subsection{Discussion}%---Learning for Interlocutor Prediction and RGMPC}
\label{Subsetion: prn-pred}
\textbf{Interlocutor Prediction and RGMPC.} 
The above methods assume that the responding speaker and target addressee are given for RGMPC. Though the speaker and the addressee could be obtained in some situations (e.g., extracted from chat logs), it is still a researchable task to interlocutor prediction. There have been some researches to predict either the responding speaker or the target addressee based on the given textual contexts or multimodal information ~\cite{DBLP:conf/interspeech/AkhtiamovSKM17,MengCIKM,Akhtiamov:addresseeSelection}. Nevertheless, in order to obtain the interaction between interlocutor prediction and RGMPC, we further design a joint model for RGMPC and interlocutor prediction. Note that both the speaker and the addressee are predicted based on textual contexts, simultaneously. Firstly, the responding speaker is predicted from contexts: %$p(a_{p}|\mathcal{C}) = \sigma([\textbf{h}_{\mathcal{C}};\textbf{h}_{L_u}^{n}]\cdot \textbf{W} \cdot \textbf{a}_{p}$, 
\begin{equation}
p(a_{res}|\mathcal{C}) = \sigma([\textbf{h}_{\mathcal{C}};\textbf{h}_{L_u}^{n}]\cdot \textbf{W} \cdot A_{res})
\end{equation}
where $\textbf{h}_{\mathcal{C}}$ is a summary contextual vector, which is max-pooled by the final interlocutor embedding matrix ($A$), and $\textbf{h}_{L_u}^{n}$ is the hidden state of the last utterance. $\textbf{W}$ is a projected matrix. $a_{res}$ and $A_{res}$ are the ID and the embedding of the responding speaker, respectively. The responding speaker is predicted by a \textit{softmax} classifier based on the embedding similarity, and the target addressee is obtained in the same way. Secondly, the predicted interlocutors replace the gold ones for the addressee memory and extracting interlocutor's embeddings from $A$. Finally, the interlocutor prediction loss is added to the response generation loss for training. Table \ref{tab:predicting interlocutor} shows the response generation performance on the situation that responding interlocutors are given and predicted. %It supports the following conclusions:
We can observe that:

\begin{table}[t]\footnotesize%\scriptsize        
    \centering
    \setlength{\tabcolsep}{3.2pt}
    \renewcommand\arraystretch{1.1}
    \begin{tabular}{p{0.85cm}<{\centering}p{1.65cm}<{\centering}p{0.80cm}<{\centering}p{1.10cm}<{\centering}p{0.90cm}<{\centering}p{0.85cm}<{\centering}}
        %\begin{tabular}{|c|c|c|c|c|}
        \hlinew{1.2pt}
        \multirow{2}[0]{*}{\textbf{Person}} & \multicolumn{1}{c}{\textbf{Speaker /}} & \multicolumn{2}{c}{\textbf{Referenced}} & \multicolumn{2}{c}{\textbf{Unreferenced}} \\ \cline{3-6}
        & \multicolumn{1}{c}{\textbf{Addressee}} & \textbf{BLEU} & \textbf{ROUGE} & \textbf{Length} & \textbf{\#Noun} \\ \hline
        Gold  & True / True & 10.63 & 8.73 & 11.34  & 1.68  \\ \hline
        \multirow{5}[0]{*}{Predict} & $\quad* \ $ / $\ * \quad$ & 9.62 & 7.88 & 11.99  & 1.44  \\ %\cline{2-7}
        & True / True & \textbf{10.05} & \textbf{8.36} & 12.04  & 1.43  \\
        & True$\,$/ $\ * \quad$ & 9.91 & 8.18 & 11.95  & 1.43  \\
        & $\quad* \ $ /$\,$True & 9.89 & 8.21 & 11.97  & 1.43  \\
        & False / False & 9.20 & 7.41 & \textbf{12.18 } & \textbf{1.47 } \\
        \hlinew{1.2pt}
    \end{tabular}%
    
    \caption{Performance on learning interlocutor prediction and RGMPC. ``True" and ``False" means right and wrong interlocutor, respectively. ``*" represents both ``True" and ``False". The correctness of the responding speaker and target addressee is segmented by ``/". For example, ``True / *" means that the responding speaker is right, and the target addressee is right or wrong.}
    \label{tab:predicting interlocutor} %  ``Data PCT(\%)" is the percentage for each type, and it is the accuracy for ``True" predictions.
\end{table}

(1) The overall performance on predicted interlocutors (``* / *" in Table \ref{tab:predicting interlocutor}) is slightly worse than the one with gold interlocutors (the first line in Table \ref{tab:predicting interlocutor}). Nevertheless, ``* / *" still outperforms the strongest baseline (VHRED in Table \ref{tab: Overall Performances}).

(2) The correctness of interlocutor prediction has a significant impact on response generation performance. It performs the best when the responding speaker and the target addressee are predicted correctly. ``False / False" (both are mispredicted) obtains the worst performance on the referenced metrics. These results demonstrate that both responding speaker and target addressee contribute to generating better responses.

(3) Surprisingly, the unreferenced metrics perform well on ``False / False". One possible reason is that the wrong interlocutors also capture rich contexts, and it generates long and meaningful responses but with a weak correlation to the gold interlocutors. Therefore, it achieves very poor performance on the referenced metrics.

\iffalse
\subsection{Case Study}

Figure~\ref{fig:case} shows an example of responses on different models for the same dialogue context. It is clearly observed that our models (marked as \textit{\textbf{bold italic}}) generate more fluent, consistent and knowledgeable (marked as \underline{underline}) responses compared to baselines. In particular, the response given by ISRED-Att-LUA ``\textit{if you want a new kernel, you can install the kernel from the kernel repo}", not only explains the \textit{reason for kernel installation} but also suggests a \textit{source of the installation}. It fully captures the context and then produces a fluent, consistent and knowledgeable response, which is semantically similar to the gold one.
\fi

\section{Related Work}
\label{Setion: Related Work}
Our work is inspired by a large number of applications utilizing recurrent encoder-decoder frameworks~\cite{cho-EtAl:2014:EMNLP2014} on NLP tasks such as machine translation~\cite{bahdanau2014neural} %,Cho2014On}
%,Luong2015Effective}
and text summarization~\cite{chopra}. Recently, many researches extend the encoder-decoder framework on response generation. HRED \cite{Serban:2016:BED:3016387.3016435} utilizes hierarchical encoder to capture the context. VHRED \cite{serban2016hierarchical} extends HRED by adding a high-dimensional latent variable for utterances. These researches demonstrate the importance of contexts on response generation. % : the first encoder models individual utterance and the second one integrates utterances into a context

Our work is also inspired by researches on multi-party chatbots. \citeauthor{dielmann2008recognition} \shortcite{dielmann2008recognition} automatically recognize dialogue acts in multi-party speech conversations. %\citeauthor{afantenos-EtAl:2015:EMNLP} \shortcite{afantenos-EtAl:2015:EMNLP} propose a discourse parsing model for multi-party chat dialogues.
Recently, some studies focus on the three elements (speaker, addressee, response) on multi-party chatbots. \citeauthor{MengCIKM} \shortcite{MengCIKM} introduce speaker classification as a surrogate task. Addressee selection is researched by \cite{Akhtiamov:addresseeSelection}. Some researches strive to the response selection \cite{ouchi-tsuboi:2016:EMNLP2016,Zhang2017Addressee}. However, the response selection heavily relies on the candidates, and it can not generate new responses in new dialogue contexts. Response generation could solve this problem. \citeauthor{li-EtAl:2016:P16-13} \shortcite{li-EtAl:2016:P16-13} learn fixed person vector for response generation. Unfortunately, it needs to be obtained from large-scale dialogue turns, which has a sparsity issue: some interlocutors have very little dialog data. Differently, our model has no such restrictions. %% learn flexible interlocutor representations from conversational contexts.% Nevertheless, current existing chatbot engines cannot properly handle a group chat with many interlocutors \cite{DBLP:journals/corr/BayserCSBCPB17}, and there is still a long way to go.
%%  \citeauthor{ijcai2018-595} \shortcite{ijcai2018-595} assign personality to response generation with explicit knowledge base (e.g., profile $<$\textit{Person, Gender, Boy}$>$). Such explicit knowledge, especially the knowledge of complex questions, is hard to be obtained and formalized.  it can not deal with unknown persons out of the training corpus. Moreover, 
\section{Conclusion}% and Future Work}

\label{Setion: Conclusion}% and Future Work}
In this study, we formalize a novel task of Response Generation for Multi-Party Chatbots (RGMPC) and propose an end-to-end model which incorporates Interlocutor-aware Contexts into Recurrent Encoder-Decoder frameworks (ICRED) for RGMPC. Specifically, we employ interactive speaker models to capture contextual interlocutor information. Moreover, we leverage an addressee memory mechanism to enrich contextual information. Furthermore, we propose to predict both the speaker and the addressee when generating responses. Finally, we construct a corpus for RGMPC. Experimental results demonstrate the ICRED remarkably outperforms strong baselines on automatic and manual evaluation metrics.% , and we find that the correctness of interlocutor prediction has a great impact on response generation Considering the importance of speakers and addressees for the addressee is fairly competitive, and it achieves 20.1\% relative improvement compared to the baseline in terms of the BLEU score.

%In the future, we are planning to extend current work as follows: (1) combining global knowledge with dialogue contexts for RGMPC; (2) exploring more interlocutor-aware representations from conversational context. % %(e.g., knowledge base)  % (1) further analyzing the architecture for multi-party chatbots;

\section*{Acknowledgments}
This work is supported by the National Natural Science Foundation of China (No.61533018), the Natural Key R\&D Program of China (No.2017YFB1002101),  the National Natural Science Foundation of China (No.61702512) and the independent research project of National Laboratory of Pattern Recognition. This work was also supported by CCF-DiDi BigData Joint Lab.

\bibliography{conll-2019}

\begin{thebibliography}{25}
\expandafter\ifx\csname natexlab\endcsname\relax\def\natexlab#1{#1}\fi

\bibitem[{Akhtiamov et~al.(2017{\natexlab{a}})Akhtiamov, Sidorov, Karpov, and
  Minker}]{DBLP:conf/interspeech/AkhtiamovSKM17}
Oleg Akhtiamov, Maxim Sidorov, Alexey~A. Karpov, and Wolfgang Minker.
  2017{\natexlab{a}}.
\newblock Speech and text analysis for multimodal addressee detection in
  human-human-computer interaction.
\newblock In \emph{Proceedings of INTERSPEECH}, pages 2521--2525.

\bibitem[{Akhtiamov et~al.(2017{\natexlab{b}})Akhtiamov, Ubskii, Feldina,
  Pugachev, Karpov, and Minker}]{Akhtiamov:addresseeSelection}
Oleg Akhtiamov, Dmitrii Ubskii, Evgeniia Feldina, Alexey Pugachev, Alexey
  Karpov, and Wolfgang Minker. 2017{\natexlab{b}}.
\newblock Are you addressing me? multimodal addressee detection in
  human-human-computer conversations.

\bibitem[{Bahdanau et~al.(2015)Bahdanau, Cho, and Bengio}]{bahdanau2014neural}
Dzmitry Bahdanau, Kyunghyun Cho, and Yoshua Bengio. 2015.
\newblock Neural machine translation by jointly learning to align and
  translate.
\newblock \emph{Proceedings of ICLR}.

\bibitem[{Ma{\'i}ra Gatti~de Bayser et~al.(2017)Ma{\'i}ra Gatti~de Bayser,
  Souza, Braz, Candello, Pinhanez, and
  Briot}]{DBLP:journals/corr/BayserCSBCPB17}
Paulo Rodrigo~Cavalin Ma{\'i}ra Gatti~de Bayser, Renan Souza, Alan Braz,
  Heloisa Candello, Claudio~S. Pinhanez, and Jean{-}Pierre Briot. 2017.
\newblock \href {http://arxiv.org/abs/1705.01214} {A hybrid architecture for
  multi-party conversational systems}.
\newblock \emph{CoRR}, abs/1705.01214.

\bibitem[{Cho et~al.(2014)Cho, van Merrienboer, Gulcehre, Bahdanau, Bougares,
  Schwenk, and Bengio}]{cho-EtAl:2014:EMNLP2014}
Kyunghyun Cho, Bart van Merrienboer, Caglar Gulcehre, Dzmitry Bahdanau, Fethi
  Bougares, Holger Schwenk, and Yoshua Bengio. 2014.
\newblock Learning phrase representations using rnn encoder--decoder for
  statistical machine translation.
\newblock In \emph{Proceedings of EMNLP}, pages 1724--1734.

\bibitem[{Chopra et~al.(2016)Chopra, Auli, and Rush}]{chopra}
Sumit Chopra, Michael Auli, and Alexander~M. Rush. 2016.
\newblock Abstractive sentence summarization with attentive recurrent neural
  networks.
\newblock In \emph{Proceedings of NAACL}, pages 93--98.

\bibitem[{Dielmann and Renals(2008)}]{dielmann2008recognition}
Alfred Dielmann and Steve Renals. 2008.
\newblock Recognition of dialogue acts in multiparty meetings using a switching
  dbn.
\newblock \emph{IEEE transactions on audio, speech, and language processing},
  pages 1303--1314.

\bibitem[{He et~al.(2017{\natexlab{a}})He, Balakrishnan, Eric, and
  Liang}]{he-EtAl:2017:Long4}
He~He, Anusha Balakrishnan, Mihail Eric, and Percy Liang. 2017{\natexlab{a}}.
\newblock Learning symmetric collaborative dialogue agents with dynamic
  knowledge graph embeddings.
\newblock In \emph{Proceedings of ACL}, pages 1766--1776.

\bibitem[{He et~al.(2017{\natexlab{b}})He, Liu, Liu, and
  Zhao}]{he-EtAl:2017:Long1}
Shizhu He, Cao Liu, Kang Liu, and Jun Zhao. 2017{\natexlab{b}}.
\newblock Generating natural answers by incorporating copying and retrieving
  mechanisms in sequence-to-sequence learning.
\newblock In \emph{Proceedings of ACL}, pages 199--208.

\bibitem[{Li et~al.(2016{\natexlab{a}})Li, Galley, Brockett, Gao, and
  Dolan}]{li-EtAl:2016:N16-11}
Jiwei Li, Michel Galley, Chris Brockett, Jianfeng Gao, and Bill Dolan.
  2016{\natexlab{a}}.
\newblock A diversity-promoting objective function for neural conversation
  models.
\newblock In \emph{Proceedings of NAACL}, pages 110--119.

\bibitem[{Li et~al.(2016{\natexlab{b}})Li, Galley, Brockett, Spithourakis, Gao,
  and Dolan}]{li-EtAl:2016:P16-13}
Jiwei Li, Michel Galley, Chris Brockett, Georgios Spithourakis, Jianfeng Gao,
  and Bill Dolan. 2016{\natexlab{b}}.
\newblock A persona-based neural conversation model.
\newblock In \emph{Proceedings of ACL}, pages 994--1003.

\bibitem[{Lin(2004)}]{Lin:2004}
Chin-Yew Lin. 2004.
\newblock Rouge: A package for automatic evaluation of summaries.
\newblock In \emph{Proceedings of ACL workshop}, page~10.

\bibitem[{Liu et~al.(2018)Liu, He, Liu, and Zhao}]{ijcai2018-587}
Cao Liu, Shizhu He, Kang Liu, and Jun Zhao. 2018.
\newblock Curriculum learning for natural answer generation.
\newblock In \emph{Proceedings of IJCAI}, pages 4223--4229.

\bibitem[{Meng et~al.(2017)Meng, Mou, and Jin}]{MengCIKM}
Zhao Meng, Lili Mou, and Zhi Jin. 2017.
\newblock Hierarchical rnn with static sentence-level attention for text-based
  speaker change detection.
\newblock In \emph{Proceedings of CIKM}, pages 2203--2206.

\bibitem[{Mou et~al.(2016)Mou, Song, Yan, Li, Zhang, and
  Jin}]{mou-EtAl:2016:COLING}
Lili Mou, Yiping Song, Rui Yan, Ge~Li, Lu~Zhang, and Zhi Jin. 2016.
\newblock Sequence to backward and forward sequences: A content-introducing
  approach to generative short-text conversation.
\newblock In \emph{Proceedings of COLING}, pages 3349--3358.

\bibitem[{Novikova et~al.(2017)Novikova, Du\v{s}ek, Cercas~Curry, and
  Rieser}]{novikova-EtAl:2017:EMNLP2017}
Jekaterina Novikova, Ond\v{r}ej Du\v{s}ek, Amanda Cercas~Curry, and Verena
  Rieser. 2017.
\newblock Why we need new evaluation metrics for nlg.
\newblock In \emph{Proceedings of EMNLP}, pages 2241--2252.

\bibitem[{Ouchi and Tsuboi(2016)}]{ouchi-tsuboi:2016:EMNLP2016}
Hiroki Ouchi and Yuta Tsuboi. 2016.
\newblock Addressee and response selection for multi-party conversation.
\newblock In \emph{Proceedings of EMNLP}, pages 2133--2143.

\bibitem[{Papineni et~al.(2002)Papineni, Roukos, Ward, and
  Zhu}]{papineni-EtAl:2002:ACL}
Kishore Papineni, Salim Roukos, Todd Ward, and Wei-Jing Zhu. 2002.
\newblock Bleu: a method for automatic evaluation of machine translation.
\newblock In \emph{Proceedings of ACL}, pages 311--318.

\bibitem[{Serban et~al.(2016)Serban, Sordoni, Bengio, Courville, and
  Pineau}]{Serban:2016:BED:3016387.3016435}
Iulian~V. Serban, Alessandro Sordoni, Yoshua Bengio, Aaron Courville, and
  Joelle Pineau. 2016.
\newblock Building end-to-end dialogue systems using generative hierarchical
  neural network models.
\newblock In \emph{Proceedings of AAAI}, pages 3776--3783.

\bibitem[{Serban et~al.(2017)Serban, Sordoni, Lowe, Charlin, Pineau, Courville,
  and Bengio}]{serban2016hierarchical}
Iulian~Vlad Serban, Alessandro Sordoni, Ryan Lowe, Laurent Charlin, Joelle
  Pineau, Aaron~C. Courville, and Yoshua Bengio. 2017.
\newblock A hierarchical latent variable encoder-decoder model for generating
  dialogues.
\newblock In \emph{Proceedings of AAAI}.

\bibitem[{Sutskever et~al.(2014)Sutskever, Vinyals, and
  Le}]{sutskever2014sequence}
Ilya Sutskever, Oriol Vinyals, and Quoc~V Le. 2014.
\newblock Sequence to sequence learning with neural networks.
\newblock In \emph{Proceedings of NIPS}, pages 3104--3112.

\bibitem[{Tian et~al.(2017)Tian, Yan, Mou, Song, Feng, and
  Zhao}]{tian-EtAl:2017:Short}
Zhiliang Tian, Rui Yan, Lili Mou, Yiping Song, Yansong Feng, and Dongyan Zhao.
  2017.
\newblock How to make context more useful? an empirical study on context-aware
  neural conversational models.
\newblock In \emph{Proceedings of ACL}, pages 231--236.

\bibitem[{Turing(1950)}]{utomated1950Computing}
Turing. 1950.
\newblock Computing machinery and intelligence.
\newblock \emph{Mind}, pages 433--460.

\bibitem[{Zhang et~al.(2018)Zhang, Lee, Polymenakos, and
  Radev}]{Zhang2017Addressee}
Rui Zhang, Honglak Lee, Lazaros Polymenakos, and Dragomir Radev. 2018.
\newblock Addressee and response selection in multi-party conversations with
  speaker interaction rnns.
\newblock In \emph{Proceedings of AAAI}.

\bibitem[{Zhou et~al.(2018)Zhou, Young, Huang, Zhao, Xu, and
  Zhu}]{ijcai2018-643}
Hao Zhou, Tom Young, Minlie Huang, Haizhou Zhao, Jingfang Xu, and Xiaoyan Zhu.
  2018.
\newblock Commonsense knowledge aware conversation generation with graph
  attention.
\newblock In \emph{Proceedings of IJCAI}, pages 4623--4629.

\end{thebibliography}
\bibliographystyle{acl_natbib}

\end{document}